%% file: main.tex
\documentclass{article}
\usepackage{spconf,amsmath,graphicx}
\usepackage{xcolor}
\usepackage{multirow}
\usepackage{caption}
\usepackage{subcaption}
\usepackage{float}
\usepackage{siunitx}
\usepackage{hyperref}
\usepackage{color, colortbl}
\usepackage{array}
\usepackage{nicematrix}

\newcommand{\etal}{\textit{et al}.}
\newcommand{\ie}{\textit{i}.\textit{e}.}
\newcommand{\eg}{\textit{e}.\textit{g}.}

\makeatletter
\newcommand{\thickhline}{%
    \noalign {\ifnum 0=`}\fi \hrule height 1pt
    \futurelet \reserved@a \@xhline
}
\newcolumntype{?}{!{\vrule width 1pt}}

\definecolor{Gray}{gray}{0.85}

\title{Diverse Generative Perturbations on Attention Space for Transferable Adversarial Attacks}
\name{Woo Jae Kim, Seunghoon Hong, and Sung-Eui Yoon
    \thanks{This work was supported by the National Research Foundation of Korea(NRF) grant funded by the Korea government(MSIT) (No. 2019R1A2C3002833, No. 2021R1A4A3032834).
}
}
\address{Korea Advanced Institute of Science and Technology (KAIST)}
\begin{document}
\ninept
\maketitle

\input{abtract}
\input{introduction}

\input{methods}

\input{experiments}

\input{conclusion}

\bibliographystyle{IEEEbib}
\bibliography{main}

\input{appendix}

\end{document}

%% file: abtract.tex
\begin{abstract}
Adversarial attacks with improved transferability -- the ability of an adversarial example crafted on a known model to also fool unknown models -- have recently received much attention due to their practicality.
Nevertheless, existing transferable attacks craft perturbations in a deterministic manner and often fail to fully explore the loss surface, thus falling into a poor local optimum and suffering from low transferability.
To solve this problem, we propose Attentive-Diversity Attack (ADA), which disrupts diverse salient features in a stochastic manner to improve transferability.
Primarily, we perturb the image attention to disrupt universal features shared by different models.
Then, to effectively avoid poor local optima, we disrupt these features in a stochastic manner and explore the search space of transferable perturbations more exhaustively.
More specifically, we use a generator to produce adversarial perturbations that each disturbs features in different ways depending on an input latent code.
Extensive experimental evaluations demonstrate the effectiveness of our method, outperforming the transferability of state-of-the-art methods. Codes are available at \url{https://github.com/wkim97/ADA}.


\end{abstract}

%% file: introduction.tex
\section{Introduction}
\label{sec:intro}

While deep neural networks (DNNs) have achieved impressive performance on numerous vision tasks~\cite{resnet, inc-v3, segmentation},
recent studies~\cite{fgsm, intriguing} have revealed their vulnerability against adversarial examples, which are crafted by adding a maliciously designed perturbation to the image.
Such adversarial attack is categorized as either white-box or black-box depending on the knowledge of the model owned by the attacker, and recent works have focused on more challenging black-box attacks.
Query-based attacks~\cite{decision-based, limited-queries, risk} use the query outputs to estimate the gradients of an unknown model, but the excessive number of required queries limits their practicality.
Instead, transfer-based attacks that rely on transferability, which is the ability of an adversarial example crafted on a white-box surrogate model to fool black-box target models, have received more attention.

\begin{figure}[!t]
    \centering
    \includegraphics[width=0.95\linewidth]{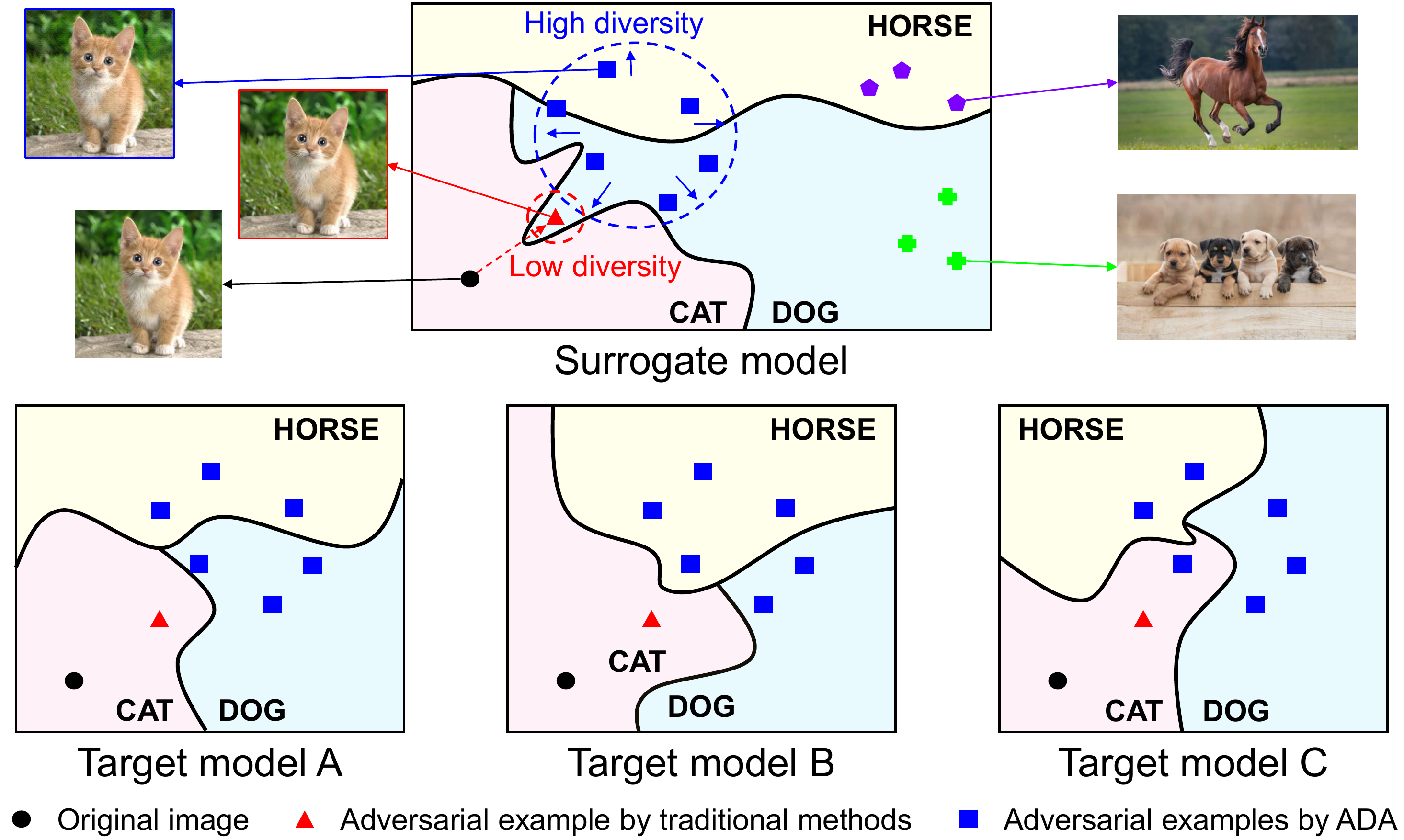}
	\vspace{-5pt}
    \caption{
    Schematic illustration of adversarial examples crafted by traditional methods and our method (ADA) along the class decision boundaries of the surrogate model and the target models.
    With lack of diversity, traditional methods greedily craft a deterministic adversarial example that easily falls into a poor local optimum and thus overfits to the surrogate model.
    In contrast, our method explores the search space of adversarial examples more exhaustively by generating diverse perturbations and avoids such local optimum.
    }
    \vspace{-15pt}
    \label{fig:fig1}
\end{figure}

However, traditional attacks (\eg, BIM~\cite{bim}, \textit{etc}.) easily overfit to the surrogate model and exhibit poor transferability.
To solve this issue, some have proposed more advanced optimization algorithms.
Dong \etal~\cite{mi-fgsm} applied momentum to avoid poor local optimum, Xie \etal~\cite{di-fgsm} applied random transformations to the image, 
Dong \etal~\cite{ti-fgsm} proposed a translation-invariant attack, 
and Wang \etal~\cite{vmi-fgsm} applied variance tuning for more stable momentum. 
Based on findings that different models learn similar features, Zhou \etal~\cite{tap} maximized the distance between the features of the original image and the adversarial image.
However, classifiers tend to also learn model-specific features~\cite{FeaturesNotBugs}, and 
more recent works perturbed salient features; Wu \etal~\cite{ata} disrupted the attention heatmaps, and Wang \etal~\cite{FIA} proposed aggregated gradients to perturb object-aware features.

Nevertheless, these attacks rely on a gradient-based method~\cite{bim} that generates perturbations in a deterministic manner.
At each iteration, they update a perturbation only in a single, specific direction that maximizes the given loss function, and with lack of stochasticity in this process, they often fail to fully explore the entire loss surface.
Thus, as shown in Fig.~\ref{fig:fig1}, with low diversity, adversarial examples crafted by these attacks easily fall into a poor local optimum and \textit{overfit} to the surrogate model, suffering from low transferability.

To solve this problem, we propose Attentive-Diversity Attack (ADA), which improves the transferability of adversarial examples by disrupting salient features in a diverse manner.
Primarily, based on recent findings~\cite{ata, FIA} that disrupting salient features boosts transferability, we perturb the image attention, which highlights features that are responsible for model decision and are likely to be shared across different models.
Greedily corrupting the attention using a gradient-based method, however, may lead to deterministic adversarial examples that easily fall into poor local optima.
To avoid such local optimum, for the first time, we propose to disrupt these features in a diverse and stochastic manner.
More specifically, we guide a generator to craft diverse perturbations that disrupt the attention differently depending on an input latent code.
As shown in Fig.~\ref{fig:fig1}, in that way, we can explore the search space of transferable adversarial examples more exhaustively, and the generator can learn to craft diverse perturbations that are located outside the poor local optimum.
These adversarial examples effectively fool the target models, while those crafted by existing deterministic methods become overfitted to the surrogate model.

In summary, our contributions are as follow:
\begin{itemize}
    
    \itemsep0em 
    
    \item For the first time, we introduce stochasticity to adversarial examples in the feature level to improve transferability.

    \item We propose Attentive-Diversity Attack (ADA), an effective generator-based adversarial attack framework that perturbs image attention in a diverse, non-deterministic manner. 
    
    
    \item Extensive experiments exhibit the superior transferability of our method as compared to existing state-of-the-art methods.
\end{itemize}

%% file: methods.tex
\begin{table*}[ht]
	\begin{center}
		\resizebox{!}{105pt}
		{\begin{tabular}{|c|c|c|c|c|c|c|c|c||c|}
		\hline
		                            \multicolumn{1}{|c}{} & \multicolumn{1}{c}{} & \multicolumn{1}{c}{} & \multicolumn{6}{|c||}{Target models} & \\
        \hline
			                        & & Attack        & Inception V3    & Inception V4    & Inception-ResNet V2 & ResNet V2    & VGG16  & Ensemble & Rank \\
		\hline
		\hline
		\multirow{24}{*}{Surrogate models} & \multirow{6}{*}{Inception V3}     & MI-FGSM       & (97.9\%)   & 42.9\%    & 39.9\%    & 41.2\%    & 53.1\%    & 35.7\%  & 6 \\
			                        & & DIM           & (98.0\%)   & 68.3\%    & 61.9\%    & 53.1\%    & 68.6\%    & 58.2\% & 5 \\
                                    & & VMI-FGSM      & (97.9\%)   & 69.6\%    & 66.7\%    & 57.6\%    & 70.0\%    & 61.8\% & 4 \\
                                    & & TAP           & (\textbf{100.0\%})  & 77.9\%    & 75.3\%    & 53.1\%    & 70.6\%    & 69.1\% & 3 \\
			                        & & FIA           & (98.5\%)   & 84.2\%    & 80.1\%    & 69.3\%    & 85.6\%    & 77.6\% & 2 \\
			                        & &\cellcolor[gray]{0.85}Ours          &\cellcolor[gray]{0.85}(96.1\%)   &\cellcolor[gray]{0.85}\textbf{88.9\%}    &\cellcolor[gray]{0.85}\textbf{82.9\%}   &\cellcolor[gray]{0.85}\textbf{82.4\%}    &\cellcolor[gray]{0.85}\textbf{95.2\%}    &\cellcolor[gray]{0.85}\textbf{85.3\%} &\cellcolor[gray]{0.85}\textbf{1} \\
		\cline{2-10}
		& \multirow{6}{*}{Inception V4}     & MI-FGSM       & 59.4\%    & (98.9\%)    & 44.9\%    & 47.8\%    & 63.6\%    & 44.9\% & 6 \\
			                        & & DIM           & 75.5\%    & (97.9\%)    & 66.5\%    & 60.5\%    & 74.9\%    & 66.5\% & 5 \\
                                    & & VMI-FGSM      & 76.6\%    & (98.5\%)    & 70.0\%    & 65.1\%    & 76.8\%    & 68.9\% & 4 \\
                                    & & TAP           & 75.6\%    & (\textbf{100.0\%})    & 70.2\%    & 59.7\%    & 82.6\%    & 72.2\% & 3 \\
			                        & & FIA           & 83.3\%    & (95.0\%)    &\textbf{78.5\%}    & 74.6\%    & 85.2\%    & 78.3\% & 2 \\
			                        & &\cellcolor[gray]{0.85}Ours          &\cellcolor[gray]{0.85}\textbf{85.2\%}   &\cellcolor[gray]{0.85}(97.7\%)    &\cellcolor[gray]{0.85}67.6\%    &\cellcolor[gray]{0.85}\textbf{79.0\%}    &\cellcolor[gray]{0.85}\textbf{89.5\%}    &\cellcolor[gray]{0.85}\textbf{79.5\%} &\cellcolor[gray]{0.85}\textbf{1} \\
        \cline{2-10}
		& \multirow{6}{*}{Inception-ResNet V2}  & MI-FGSM   & 58.0\%    & 52.6\%    &(\textbf{99.4\%})    & 46.9\%    & 63.5\%    & 47.0\% & 6 \\
			                        & & DIM           & 73.0\%    & 70.6\%    & (94.8\%)    & 57.7\%    & 70.4\%    & 65.6\% & 4 \\
                                    & & VMI-FGSM      & 78.4\%    & 77.4\%    & (99.3\%)    & 64.5\%    & 74.2\%   & 71.6\% & 3  \\
                                    & & TAP           & 74.1\%    & 66.8\%    & (95.2\%)    & 46.7\%    & 63.0\%    & 51.8\% & 5 \\
			                        & & FIA           & 81.6\%    & 77.1\%    & (88.6\%)    & 66.9\%    & 81.9\%    & 74.8\% & 2 \\
			                        & &\cellcolor[gray]{0.85}Ours          &\cellcolor[gray]{0.85}\textbf{82.5\%}    &\cellcolor[gray]{0.85}\textbf{89.4\%}    &\cellcolor[gray]{0.85}(93.0\%)    &\cellcolor[gray]{0.85}\textbf{80.7\%}    &\cellcolor[gray]{0.85}\textbf{89.4\%}  &\cellcolor[gray]{0.85}\textbf{85.9\%} &\cellcolor[gray]{0.85}\textbf{1} \\
        \cline{2-10}                        
		& \multirow{6}{*}{ResNet V2}     & MI-FGSM       & 54.7\%    & 47.9\%    & 44.1\%    & (99.6\%)    & 61.4\%    & 44.4\% & 6 \\
			                        & & DIM           & 75.3\%    & 70.6\%    & 68.8\%    & (99.1\%)    & 73.9\%    & 70.0\% & 3 \\
                                    & & VMI-FGSM      & 72.7\%    & 67.4\%    & 64.7\%    & (97.6\%)    & 72.3\%    & 65.3\% & 4 \\
                                    & & TAP           & 51.8\%    & 44.3\%    & 44.5\%    & (92.4\%)    & 68.2\%    & 52.6\% & 5 \\
			                        & & FIA           &\textbf{81.2\%}    & 76.7\%    &\textbf{74.5\%}   &(\textbf{99.9\%})   & 82.9\%    & 74.1\% & 2 \\
			                        & &\cellcolor[gray]{0.85}Ours          &\cellcolor[gray]{0.85}79.7\%    &\cellcolor[gray]{0.85}\textbf{90.9\%}    &\cellcolor[gray]{0.85}71.9\%    &\cellcolor[gray]{0.85}(94.0\%)    &\cellcolor[gray]{0.85}\textbf{93.1\%}    &\cellcolor[gray]{0.85}\textbf{80.6\%} &\cellcolor[gray]{0.85}\textbf{1} \\
        \hline              
        \end{tabular}}
	\end{center}
	\vspace{-15pt}
	\caption{Attack success rates of different attacks against various target models. The leftmost column and the uppermost row show surrogate models and target models, respectively. Parentheses () indicates white-box attack where the target model is the surrogate model. Best results are highlighted in bold, and ``Rank" denotes the order of highest average ASR on black-box models.}
	\label{table:tab1}
	\vspace{-15pt}
\end{table*}

\section{Methods}
\label{sec:methods}

\vspace{0.5ex}\noindent
\textbf{Preliminaries.}
Let $f_\psi$ be a target classifier. 
The objective of an untargeted adversarial attack is to create an adversary $x^{adv}$ of an image $x$ in class $t$ such that it leads to a misclassification on the target classifier (\ie, $f_\psi(x^{adv}) \neq t$). 
In this paper, we consider a black-box attack where we do not have access to the target classifier. 
Instead, we employ an accessible surrogate model $h_\theta$ that shares the same output space with $f_\psi$ but has different architectures and/or parameters.
We then generate a transferable adversarial example on the surrogate model as follows:
\begin{equation}
    \vspace{-2pt}
    \arg \max_{x^{adv}} L_\theta(x^{adv}, t), \quad \text{s.t.} \quad \|x - x^{adv}\|_\infty \leq \epsilon,
    \label{eq:eq1}
    \vspace{-2pt}
\end{equation}
where $L_\theta(\cdot, \cdot)$ is the classification loss on the surrogate model $h_\theta$, and $\epsilon$ is a constraint set on the magnitude of perturbation.

The success of such transfer-based attack highly depends on its transferability.
Nevertheless, existing attempts rely on a gradient-based method to optimize Eq.~\ref{eq:eq1} and thus greedily find a deterministic solution that maximizes $L_\theta(x^{adv}, t)$.
Such deterministic property can overfit the adversarial examples to the surrogate model and thus lower the chance of black-box attack being successful.

\begin{figure}[!t]
    \centering
    \includegraphics[width=0.9\linewidth]{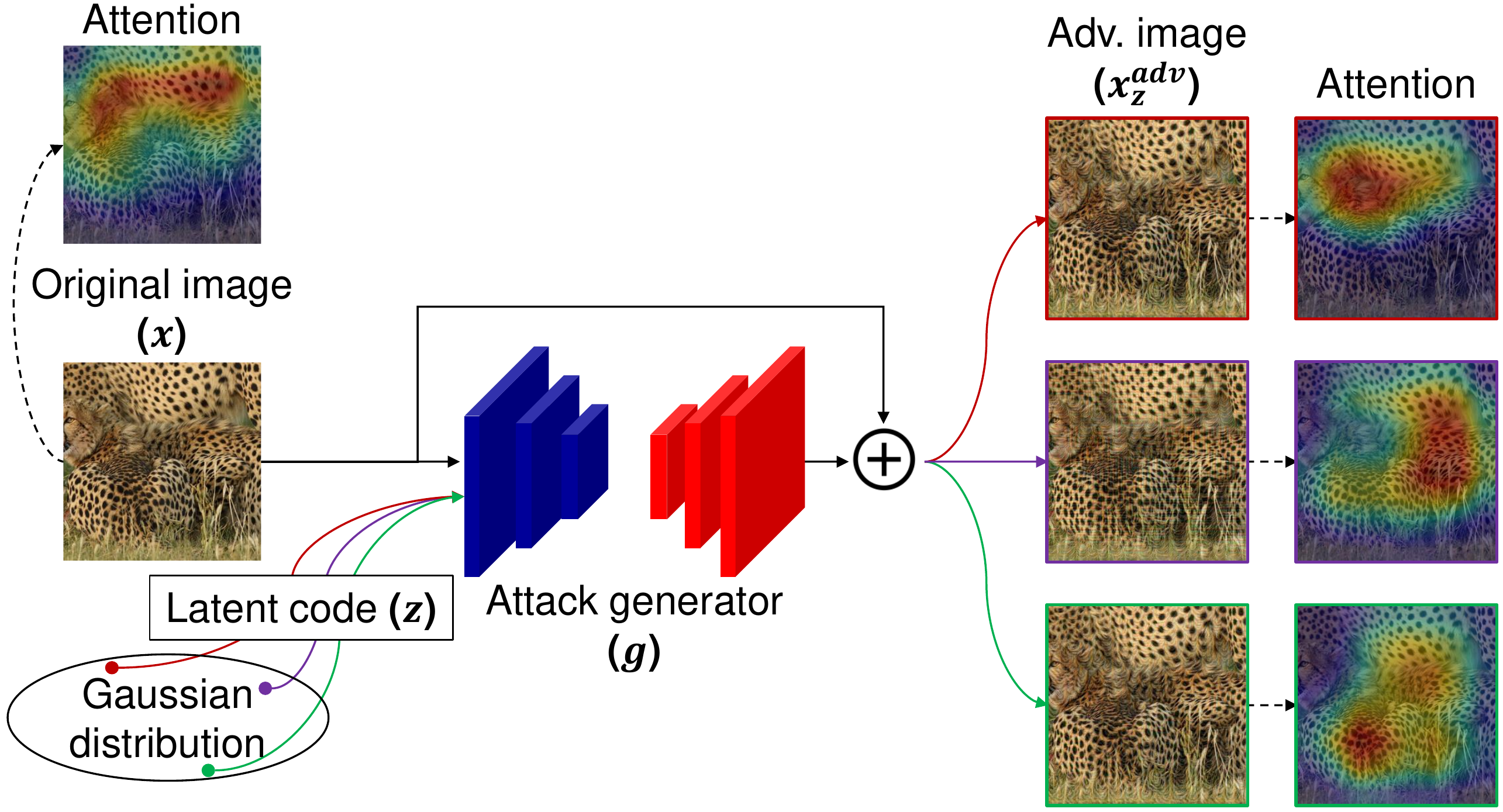}
    \vspace{-5pt}
    \caption{Overview of Attentive-Diversity Attack (ADA). Given an image and a latent code, the attack generator produces a perturbation that disrupts the image attention in a diverse manner.}
    \vspace{-15pt}
    \label{fig:fig2}
\end{figure}

To address this challenge, we propose Attentive-Diversity Attack (ADA) (Fig.~\ref{fig:fig2}), which significantly improves the transferability of adversarial examples by using an \textit{attack generator} guided by \textit{attention perturbation} and \textit{feature diversification}.

\vspace{0.5ex}\noindent
\textbf{Attack Generator.}
Instead of a gradient-based method that crafts perturbations in an iterative and deterministic manner, 
we use a generator to parameterize the adversary with a DNN.
Given an image $x$ and a latent code $z$ sampled from a Gaussian distribution, the generator $g$ learns to output an adversarial perturbation that is dependent on the latent code.
We then form an adversarial image $x^{adv}_z$ as follows: 
\begin{equation}
    \vspace{-2pt}
   x^{adv}_z = Clip_{x, \epsilon}\{x + \epsilon \cdot g(x, z)\},
   \label{eq:eq2}
    \vspace{-2pt}
\end{equation}
where $Clip_{x, \epsilon}$ clips the perturbation in a per-pixel manner so that it is bounded to $\epsilon$-ball of $L_\infty$ norm.


\vspace{0.5ex}\noindent
\textbf{Attention Perturbation.}
In order to boost transferability, we disrupt the image attention, which highlights features that are responsible for model decision and are likely to be relevant to the main objects of the image.
As different classifiers universally rely on these object-related features to make decisions, perturbations on these features will effectively transfer to other models.


Based on Grad-CAM~\cite{gradcam}, we define attention $A$ as a weighted representation of features $F$, which we set as the output from the last convolutional layer (\eg, \textit{Mixed\_7c} for Inc-v3), as follows:
\begin{equation}
    \vspace{-2pt}
    A(x; t) = \alpha_t F = \texttt{GAP} \left( \frac{\partial y_t}{\partial F} \right) F.
    \label{eq:eq3}
    \vspace{-2pt}
\end{equation}
The weight $\alpha_t$ denotes the importance of the feature $F$ given the ground truth class $t$.
It is obtained by taking the gradient of $y_t$ -- the prediction for class $t$ -- with respect to $F$ and applying global average pooling ($\texttt{GAP}(\cdot)$) over the spatial dimension.
To prevent the generator from perturbing only the few channels with the highest magnitudes, we further apply channel-wise normalization on $A(x; t)$. 

Then, the generator learns to maximize the distance between the attention representations of the original image and the adversarial image by maximizing the following \textit{attention loss} $L_{attn}$:
\begin{equation}
    \vspace{-2pt}
    L_{attn} = \| A(x^{adv}_z; t) - A(x; t) \|_2.
    \label{eq:eq4}
    \vspace{-2pt}
\end{equation}

While Wu \etal~\cite{ata} have similarly disrupted the attention heatmaps extracted using the techniques of Grad-CAM~\cite{gradcam}, our method differs from their approach on that we additionally apply channel normalization.
Without channel normalization, the generator perturbs only the few feature channels with highest magnitudes, which limits the diversity of perturbations it can generate.
To prevent this, we normalize each feature channel and enable the generator to disrupt more diverse features.


\vspace{0.5ex}\noindent
\textbf{Feature Diversification.}
Without any guidance, the generator may learn to ignore the input latent code and greedily maximize the attention loss in a deterministic manner just like gradient-based methods.
Thus, we guide the generator to explore and corrupt diverse features in a stochastic manner.
We train it to disturb the attention representations differently for two distinct input latent codes $z_1$ and $z_2$, each sampled from a Gaussian distribution, by applying a diversity regularization~\cite{dsgan} and maximizing the following \textit{diversity loss} $L_{div}$:
\begin{equation}
    \vspace{-2pt}
    L_{div} = \frac{\| A(x^{adv}_{z_1}; t) - A(x^{adv}_{z_2}; t) \|}{\| z_1 - z_2 \|}.
    \label{eq:eq5}
    \vspace{-2pt}
\end{equation}
We craft two adversarial examples $x^{adv}_{z_1}$ and $x^{adv}_{z_2}$ each by passing $z_1$ and $z_2$, respectively, into the generator (Eq.~\ref{eq:eq2}) and obtain their respective attention representations $A(x^{adv}_{z_1}; t)$ and $A(x^{adv}_{z_2}; t)$ (Eq.~\ref{eq:eq3}).
Then, by maximizing the distance between the two representations, we force the generator to craft semantically diverse perturbations.

While Yang \etal~\cite{dsgan} originally proposed the diversity regularization, their applications have been limited to pixel or feature levels.
Diversity on these levels, however, may not necessarily translate to diversity on the attention space and may fail to guide our generator to disrupt the \textit{salient} features in a diverse manner.
To explicitly guide it to perturb the meaningful features in a diverse manner, unlike existing approaches, we apply diversity on the attention level.

Overall, we learn the attack generator $g$ to \textit{maximize}:
\begin{equation}
    \vspace{-2pt}
    L = L_{cls} + \lambda_{attn} \cdot L_{attn} + \lambda_{div} \cdot L_{div},
    \vspace{-2pt}
\end{equation}
where $L_{cls}$ is the cross-entropy loss between the adversarial image and the ground truth label, and $\lambda_{attn}$ and $\lambda_{div}$ control the weights of the attention loss ($L_{attn}$) and the diversity loss ($L_{div}$), respectively.

There have been several attempts to craft diverse adversarial examples.
Jang \etal~\cite{l2l-da} and Dong \etal~\cite{adt} modeled diverse perturbations from a single image, but their approaches are limited to pixel-level diversity and improving adversarial robustness.
Xie \etal~\cite{di-fgsm} boosted transferability by crafting perturbations on randomly transformed images, but their approach can only implicitly perturb features in a diverse manner as a result of pixel-level transformations.
In contrast, for the first time, we craft semantically diverse perturbations by explicitly disrupting diverse features.
As a result, we effectively avoid poor local optimum, improving transferability as also shown by the experiment results (Table~\ref{table:tab1}).

%% file: experiments.tex
\begin{figure*}[t!]
    \centering
    \begin{subfigure}[t]{0.396\textwidth}
        \centering
        \includegraphics[width=\linewidth]{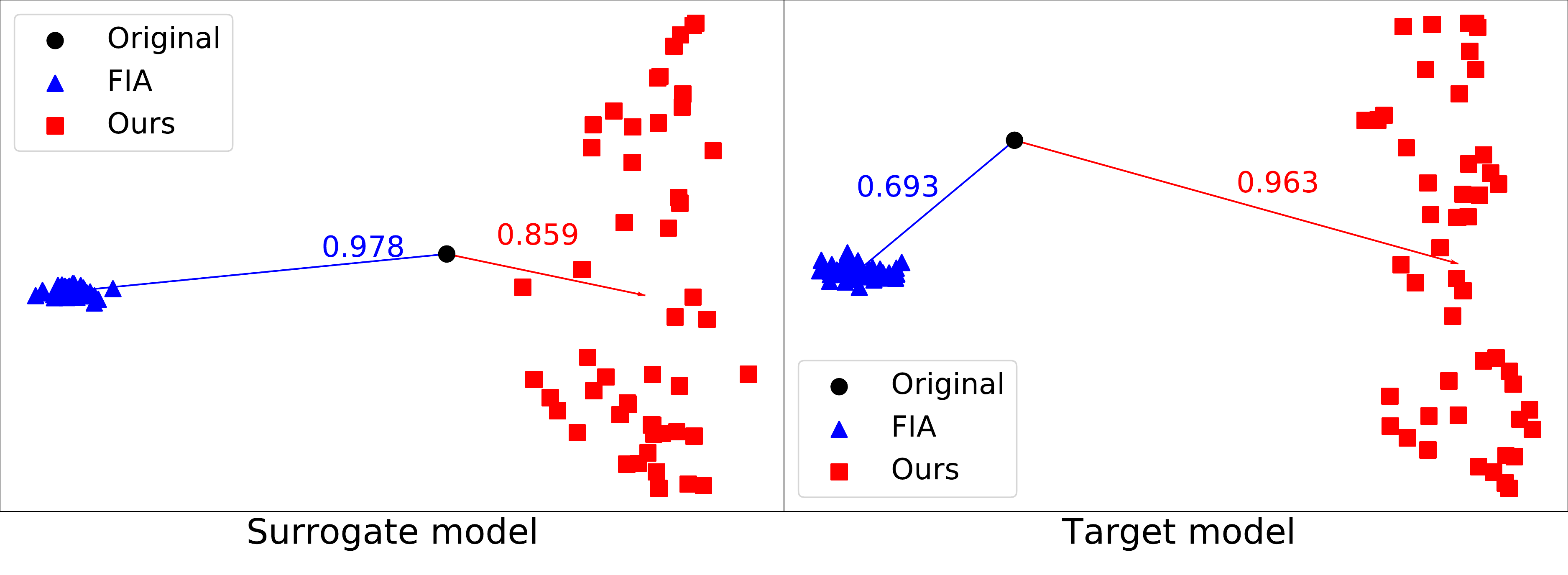}
        \caption{Distribution of adversarial examples}
        \label{fig:fig3a}
    \end{subfigure}%
    ~
    \begin{subfigure}[t]{0.18\textwidth}
        \centering
        \includegraphics[width=\linewidth]{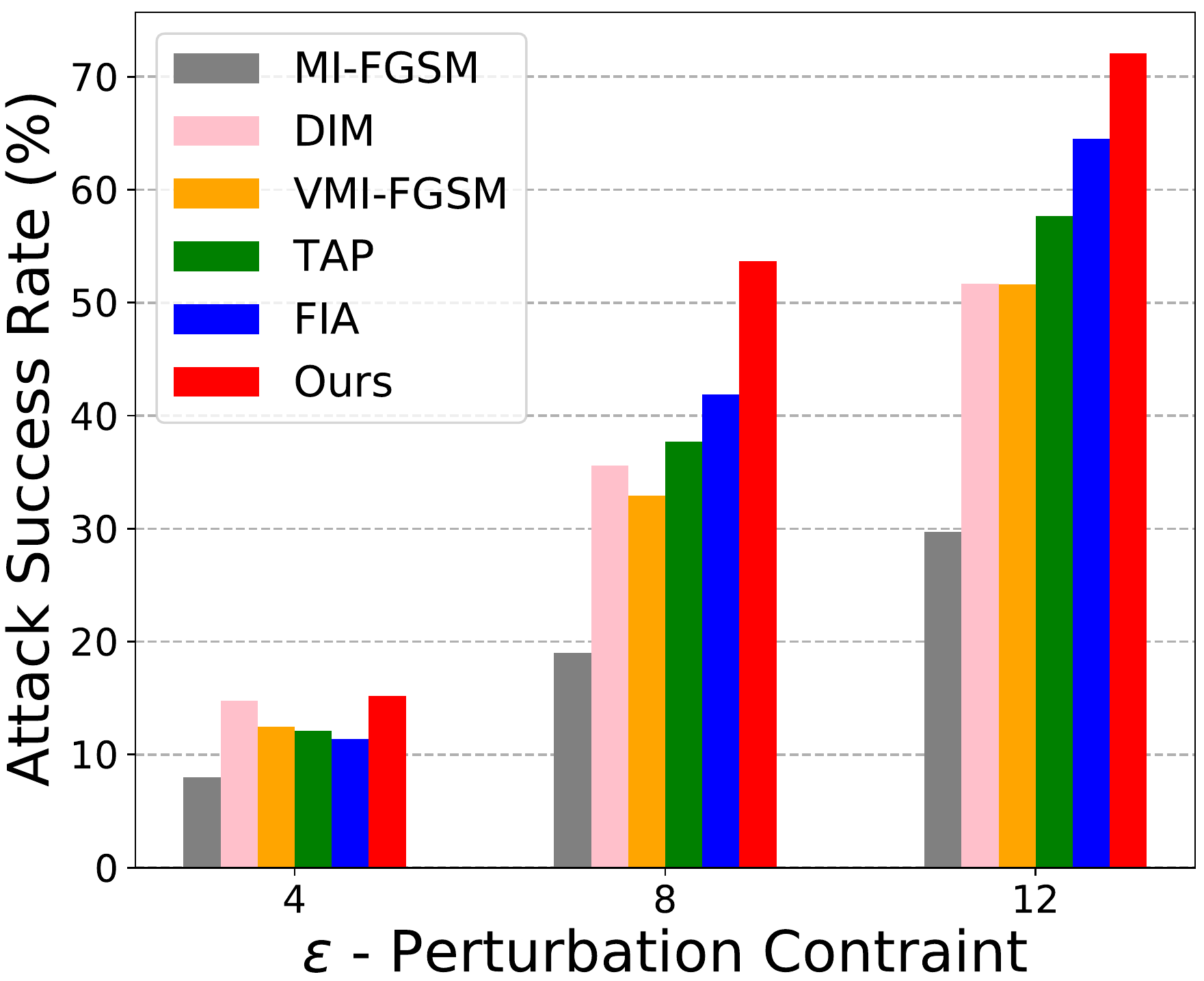}
        \caption{Comparison on various $\epsilon$}
        \label{fig:fig3b}
    \end{subfigure}%
    ~
    \begin{subfigure}[t]{0.1825\textwidth}
        \centering
        \includegraphics[width=\linewidth]{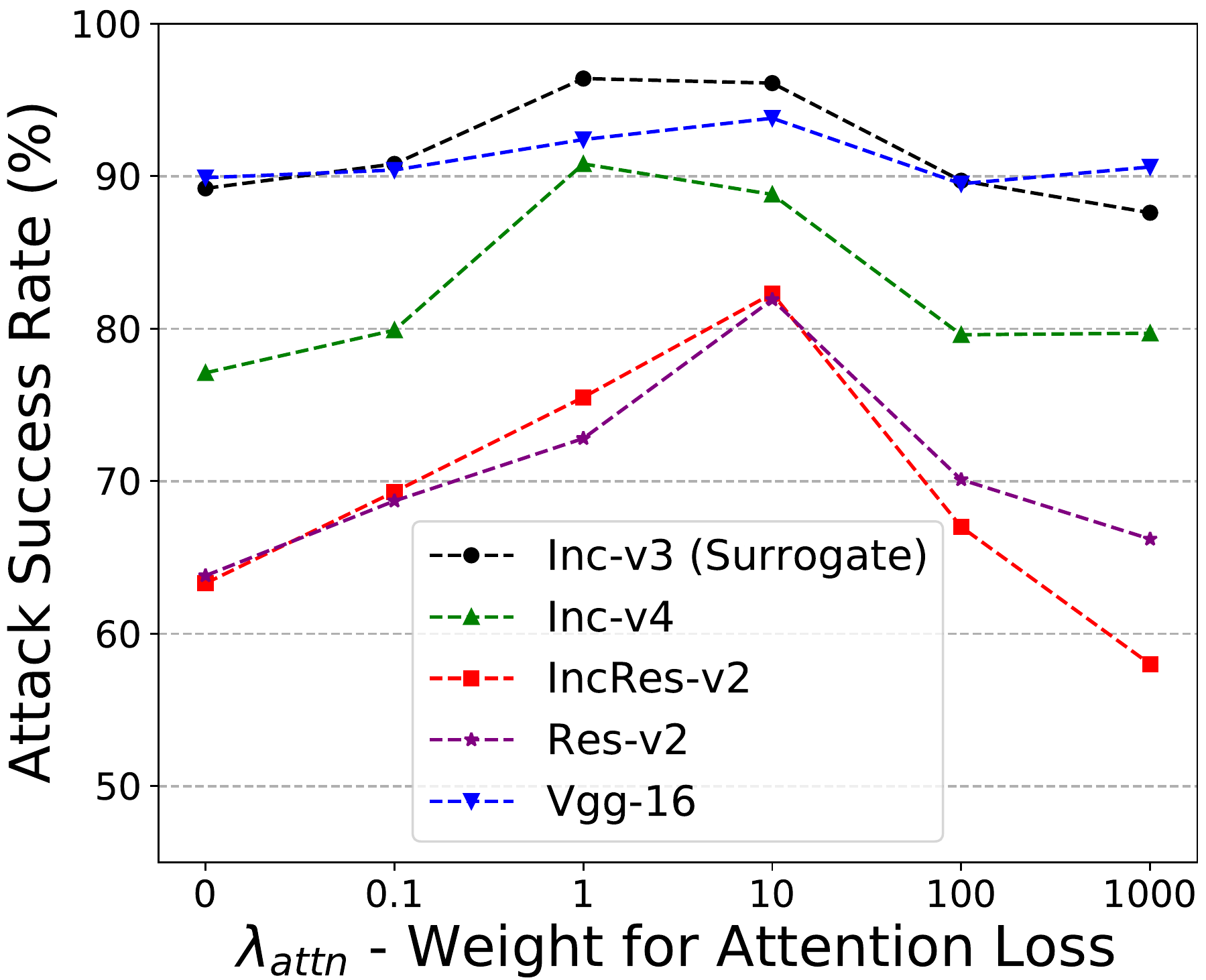}
        \caption{Effects of $L_{attn}$}
        \label{fig:fig3c}
    \end{subfigure}%
    ~
    \begin{subfigure}[t]{0.18\textwidth}
        \centering
        \includegraphics[width=\linewidth]{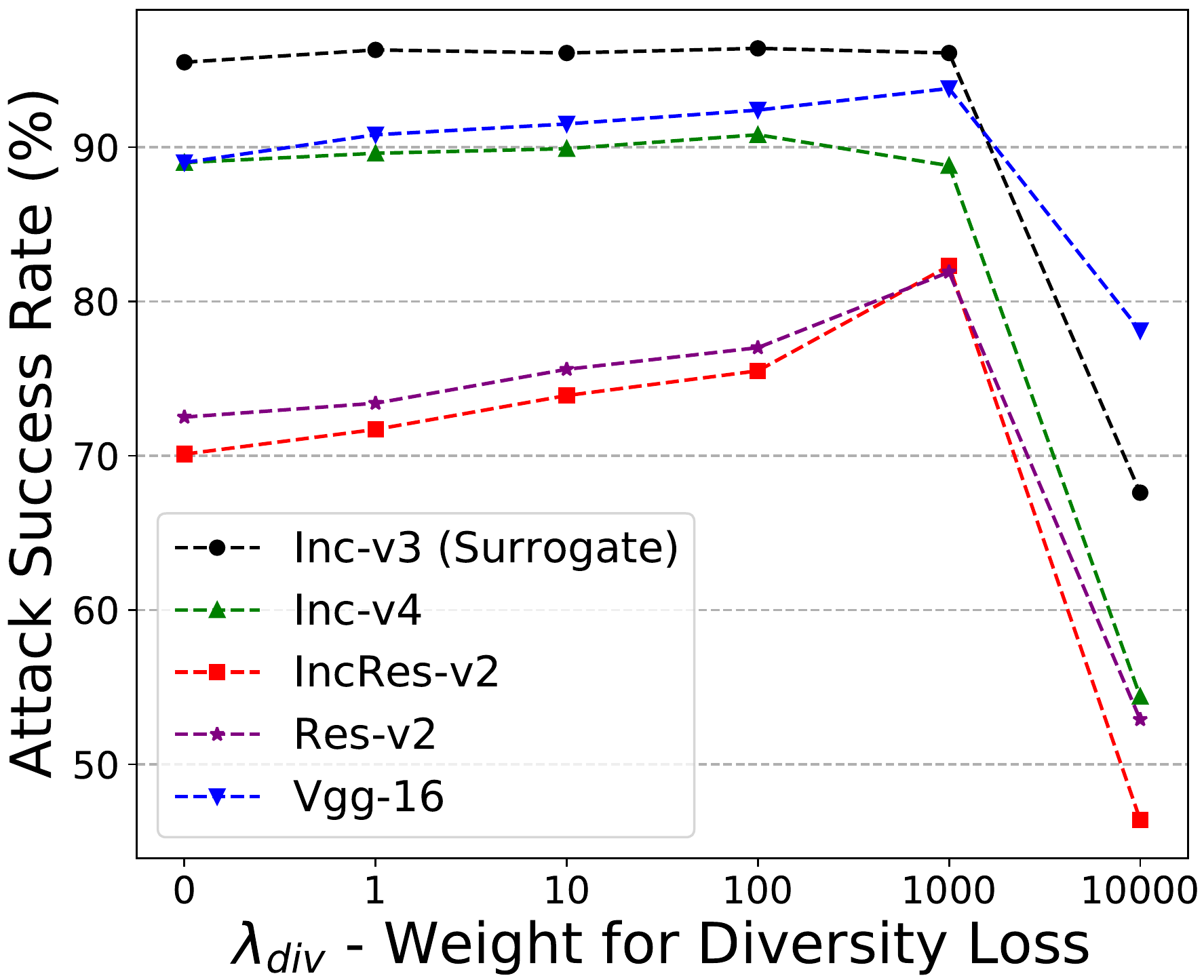}
        \caption{Effects of $L_{div}$}
        \label{fig:fig3d}
    \end{subfigure}%
    
    \vspace{-5pt}
    \caption{
    Analysis of our method with all adversarial examples crafted on Inc-v3 as the surrogate model.
    (a) PCA visualization on surrogate model (Inc-v3) and target model (Res-v2) for features of adversarial examples crafted by FIA (blue) and our method (red),
    (b) comparison of attack success rates (ASR) on the ensemble of black-box target models with varying perturbation constraint $\epsilon$, 
    (c) ASR on varying weights for attention loss $\lambda_{attn}$ when $\lambda_{div} = 1000$, and
    (d) ASR on varying weights for diversity loss $\lambda_{div}$ when $\lambda_{attn} = 10$.
    }
    \label{fig:fig3}    
    \vspace{-10pt}
\end{figure*}

\section{Experiments}

\vspace{0.5ex}\noindent
\textbf{Experiment Setup.}
For the attack generator, we employ a U-Net~\cite{gap, unet} based convolutional encoder-decoder consisting of three encoding blocks and three decoding blocks.
Each encoding and decoding block consists of a convolutional layer and a transposed convolutional layer, respectively, followed by a batch normalization layer and a ReLU layer.
At each encoding block, the latent code $z$ is spatially expanded and concatenated to the input of the block.
The generator is trained for 100 epochs with learning rate of 1e-4, batch size of 8, and an Adam optimizer~\cite{adam} with $\beta_1 = 0.5$, $\beta_2 = 0.999$, and weight decay 1e-5.

We use 10,000 images randomly selected from the ImageNet validation set~\cite{imagenet} for train data and 1,000 images from the NeurIPS 2017 adversarial competition~\cite{nips2017} for test data. 
We test our method on Inception-v3 (Inc-v3)~\cite{inc-v3}, Inception-v4 (Inc-v4)~\cite{inc-v4}, Inception-ResNet-V2 (IncRes-v2)~\cite{inc-v4}, ResNet-V2 (Res-v2)~\cite{resnetv2}, and VGG16 (Vgg-16)~\cite{vgg}\footnote{These models are publicly available at: \url{https://github.com/Cadene/pretrained-models.pytorch}}. 
We compare our method with various state-of-the-art attacks -- MI-FGSM~\cite{mi-fgsm}, DIM~\cite{di-fgsm}, VMI-FGSM~\cite{vmi-fgsm}, TAP~\cite{tap}, and FIA~\cite{FIA} -- for which we set the number of iterations $T = 10$, the step size $\alpha = 1.6$, and the rest of the hyperparameters as specified in their respective references.
The maximum perturbation constraint $\epsilon$ is set to $16$ under $L_\infty$ norm.
For our method, we use $\lambda_{attn} = 10$, $\lambda_{div} = 1000$, and 16 for the length of latent code $z$.

\vspace{0.5ex}\noindent
\textbf{Comparison of Transferability.}
To compare the transferability of our method and the baselines, we craft adversarial examples on four surrogate models -- Inc-v3, Inc-v4, IncRes-v2, and Res-v2 -- and measure their attack success rates (ASR), or the misclassification rate of a model~\cite{vmi-fgsm}, on five target models (Table~\ref{table:tab1}).
The \textbf{ensemble} denotes ASR on the ensemble of all target models excluding the white-box model (\ie, target model is different from the surrogate model), whose prediction is the average probability output of all models.
The results indicate that our method fools black-box target models with higher ASR than existing methods in most cases, outperforming FIA by average of 6.4\%p on the ensemble model.
Also, while our method generally shows lower ASR on the white-box target models, it shows much higher ASR on black-box target models, showing that it overfits less to the surrogate model and generalizes well to unknown models.


An interesting observation is that our method outperforms existing methods by a larger margin when the architectures of the target models are more different from that of the surrogate model.
For example, while our attack crafted on Inc-v3 outperforms FIA by 4.7\%p on a more similarly structured Inc-v4~\cite{comprehensive}, it outperforms FIA by a larger margin of 9.6\%p on a more differently structured Vgg-16~\cite{comprehensive}. 
While existing methods overfit to the surrogate model and show low transferability on models with more distinct structures, our method shows high transferability regardless of the model structure.


\begin{table}
	\begin{center}
	    \resizebox{0.9\columnwidth}{!}
		{\begin{tabular}{c|c|c|c|c|c}
		\hline
		& No Attack & FGSM & BIM & PGD & Ours \\
		\hline
		\hline
		No defense$^\dagger$ & 91.85\% & 35.51\% & 2.89\% & 2.59\% & 4.45\% \\
        \thickhline
        FGSM & 79.24\% & 82.25\% & 81.02\% & 83.99\% & 83.77\% \\
		BIM & 83.17\% & 82.80\% & 83.17\% & 83.10\% & 82.28\% \\
		PGD & 85.84\% & 84.73\% & 85.62\% & 85.54\% & 85.10\% \\
		\cellcolor[gray]{0.85}Ours (w/o $L_{div}$) &\cellcolor[gray]{0.85}89.18\% &\cellcolor[gray]{0.85}80.80\% &\cellcolor[gray]{0.85}85.62\% &\cellcolor[gray]{0.85}86.66\% &\cellcolor[gray]{0.85}86.95\% \\
		\cellcolor[gray]{0.85}Ours (w/ $L_{div}$) &\cellcolor[gray]{0.85}\textbf{90.88\%} &\cellcolor[gray]{0.85}\textbf{85.92\%} &\cellcolor[gray]{0.85}\textbf{88.21\%} &\cellcolor[gray]{0.85}\textbf{89.25\%} &\cellcolor[gray]{0.85}\textbf{89.62\%} \\
		
		\hline
        \end{tabular}}
	\end{center}
	\vspace{-15pt}
	\caption{
	Classification accuracy of adversarial training models under various attacks.
	Leftmost column and uppermost row represent the attacks used for training and the attacks used for evaluation, respectively. Adversarial examples used for evaluation are crafted on a classifier trained with original images (marked $\dagger$). Best results are highlighted in bold.}
	\label{table:tab2}
	\vspace{-15pt}
\end{table}

\vspace{0.5ex}\noindent
\textbf{Generation of Diverse Perturbations.}
Recent findings~\cite{l2l-da, adt} have shown that the robustness of adversarial training model improves when trained against diverse adversarial examples.
Based on this idea, we test the robustness of a classifier trained against adversarial examples crafted by ADA to show that our method indeed generates diverse perturbations.
We train Inc-v3 for 30 epochs using batch size of 8, an SGD optimizer with learning rate of 0.001, momentum of 0.9, and weight decay of 5e-4 on Caltech101 dataset~\cite{caltech101} with 8,681 images and 101 classes randomly split into 7,332/1,349 for training/test set.


Table~\ref{table:tab2} exhibits the classification accuracy of each classifier adversarially trained by a different attack.
The classifier trained by our method with $L_{div}$ records the highest robustness in all cases, outperforming that trained without $L_{div}$ by 3.24\%p in average.
While our method without $L_{div}$ and gradient-based methods (FGSM~\cite{fgsm}, BIM~\cite{bim}, and PGD~\cite{pgd}) train the classifiers only against a limited set of deterministic attacks, our method with $L_{div}$ trains the classifier against diverse adversarial examples and makes it more robust.

\begin{figure}[t!]
    \centering
    \includegraphics[width=1.0\linewidth]{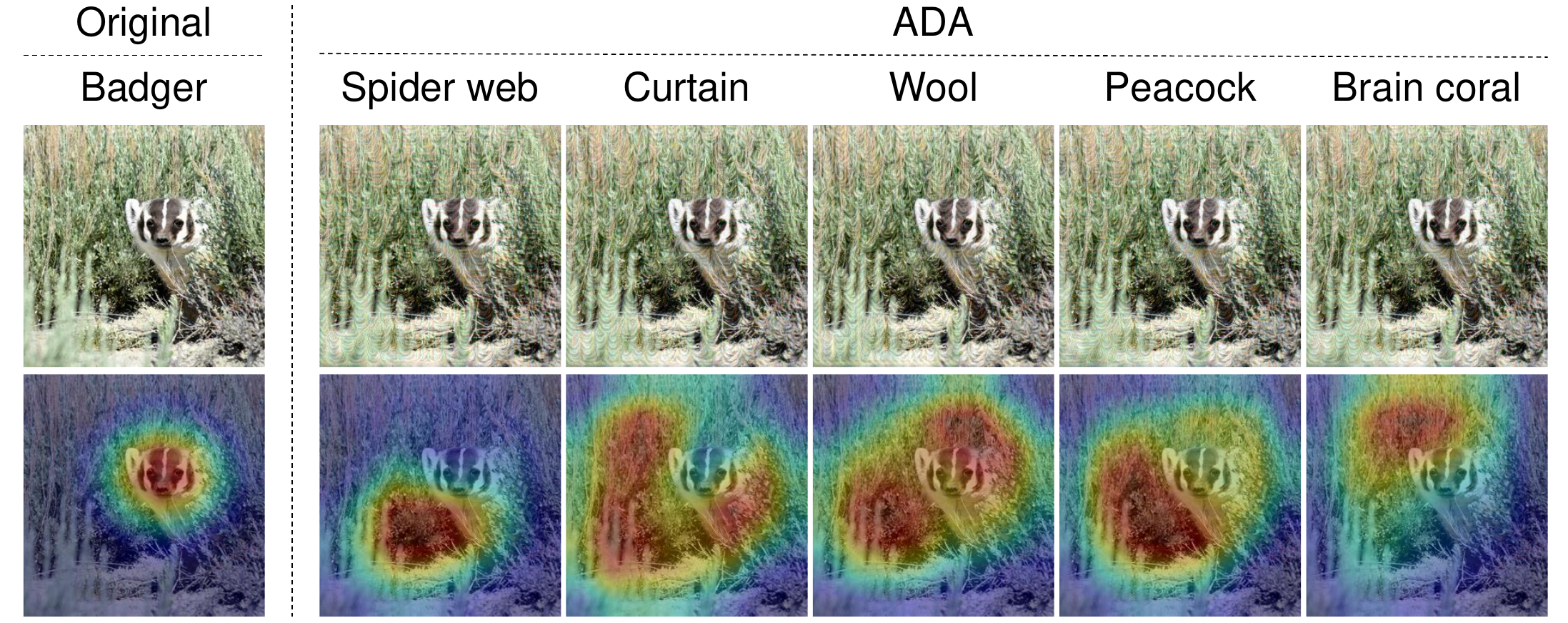}
	\vspace{-15pt}
    \caption{Original/adversarial images (top row) crafted by our attack and their attention (bottom row). Our attack crafts adversarial examples that disrupt the attention and final predictions in diverse ways.}
    \label{fig:fig4}
    \vspace{-15pt}
\end{figure}

We additionally visualize semantically diverse adversarial examples crafted by our method in Fig.~\ref{fig:fig4}.
From a single image, our method generates various adversarial examples that disrupt the image attention and model predictions in a diverse manner.

\vspace{0.5ex}\noindent
\textbf{Effects of Diversity on Transferability.}
We additionally perform a set of experiments to further analyze our attack and visualize the results in Fig.~\ref{fig:fig3}.
To prove that disrupting diverse features improves transferability, we visualize in Fig.~\ref{fig:fig3a} the features of adversarial examples each crafted by FIA and our method by projecting them on the 2D space spanned by eigenvectors obtained from PCA.
As shown in the figure, FIA generates deterministic adversarial examples whose feature representations form a dense cluster, while those of our method are widely spread out.
Also, the distribution on the surrogate model (left) shows that FIA generates stronger adversarial examples compared to our method; the average $\ell_2$ distance from the feature of the original image to the features of adversarial examples crafted by FIA is 0.978, which is higher than that of our method, which is 0.859.
However, the adversarial examples formed by FIA suffer from poor transferability on the target model, showing a far lower average distance of 0.693 as compared to 0.963 of our method.
By learning to craft semantically diverse perturbations, our method effectively avoids poor local optimum and improves transferability.

\vspace{0.5ex}\noindent
\textbf{Various Perturbation Constraint $\epsilon$.}
We also compare the transferability of our attack with the baselines under lower perturbation constraints $\epsilon$ (\ie, 4, 8, and 12) in Fig.~\ref{fig:fig3b}, which shows the ASR on the ensemble of black-box target models.
Our method exhibits superior transferability over all of the existing methods under all constraints, outperforming FIA by average ASR of 7.7\%p.
Even when perturbations are less visible, our method exhibits high transferability.

\vspace{0.5ex}\noindent
\textbf{Effects of $L_{attn}$.}
We show in Fig.~\ref{fig:fig3c} the effects of the weight $\lambda_{attn}$, which is used to control the attention loss $L_{attn}$.
In general, $L_{attn}$ plays an important role at boosting the transferability, achieving the highest performance when $\lambda_{attn} = 10$.
Interestingly, using a slightly lower weight of $\lambda_{attn} = 1$ improves ASR on Inc-v4 but worsens ASR on other black-box models. Insufficient weight on attention loss overfits the attack to the surrogate model and boosts transferability on similarly structured Inc-v4 but reduces transferability on more differently structured black-box models.
Too much focus on $L_{attn}$ disturbs the attack from disrupting the prediction output of the classifier, dropping ASR on all models.



\vspace{0.5ex}\noindent
\textbf{Effects of $L_{div}$.}
In Fig.~\ref{fig:fig3d}, we show the effects of the weight $\lambda_{div}$ that is used to control the diversity loss $L_{div}$.
Using $L_{div}$ generally improves transferability and achieves the highest performance when $\lambda_{div} = 1000$, showing that disrupting diverse features prevents the adversarial examples from overfitting to the surrogate model.
Also in this case, using a lower $\lambda_{div} = 100$ improves ASR on Inc-v4 but significantly hurts ASR on other target models, proving that the lack of diversification overfits the generator to similarly structured Inc-v3 and Inc-v4.
Too much weight on diversity loss prevents the generator from disrupting features in a destructive manner and lowers ASR.


%% file: conclusion.tex
\section{Conclusion}
In this paper, we have proposed Attentive-Diversity Attack (ADA) that generates highly transferable adversarial examples.
ADA relies on a generator to generate perturbations that disrupt image-salient features in a non-deterministic manner.
Consequently, the crafted adversarial examples avoid falling into poor local optima and become less overfitted to the surrogate model.
Exhaustive experiments validate the superior performance of ADA against state-of-the-art methods and the effectiveness of its individual components.

%% file: appendix.tex
\renewcommand\thesection{\Alph{section}}
\renewcommand\thesubsection{\thesection.\arabic{subsection}}

\addcontentsline{toc}{chapter}{First unnumbered chapter}
\setcounter{section}{0}
\renewcommand*{\theHsection}{chX.\the\value{section}}


\begin{table*}[t!]
	\begin{center}
        \resizebox{0.65\linewidth}{!}		
        {\begin{tabular}{c|c|c|c|c|c|c}
        \hline
        Attack                  & Inc-v3            & Inc-v4            & IncRes-v2         & Res-v2            & Vgg16         & Ensemble  \\
		\hline
		\hline
		Ours (w/o channel-norm) & 79.3\%            & 87.2\%            & 92.9\%            & 72.0\%            & 88.9\%            & 82.9\%    \\
		Ours                    & \textbf{85.2\%}   & \textbf{89.4\%}   & \textbf{93.0\%}   & \textbf{80.7\%}   & \textbf{89.4\%}   & \textbf{85.9\%} \\
		\hline              
        \end{tabular}}
	\end{center}
	\vspace{-15pt}
	\caption{Comparison of attack success rates as we remove channel-wise normalization from the attention loss $L_{attn}$.
	All adversarial examples are crafted using IncRes-v2 as the surrogate model.  
	Best results are highlighted in \textbf{bold}.}
	\label{table:ap-tab1}
\end{table*}

\begin{table*}[t!]
	\begin{center}
        \resizebox{0.575\linewidth}{!}		
        {\begin{tabular}{c|c|c|c|c|c|c}
        \hline
        Attack          & Inc-v3            & Inc-v4            & IncRes-v2         & Res-v2            & Vgg16             & Ensemble \\
		\hline
		\hline
		Pixel-level     & 76.3\%            & 79.0\%            & 91.7\%            & 73.8\%            & 88.2\%            & 74.8\% \\
		Feature-level   & 76.5\%            & 85.2\%            & 92.4\%            & 76.7\%            & 88.4\%            & 82.5\% \\
		Ours            & \textbf{85.2\%}   & \textbf{89.4\%}   & \textbf{93.0\%}   & \textbf{80.7\%}   & \textbf{89.4\%}   & \textbf{85.9\%} \\
		\hline              
        \end{tabular}}
	\end{center}
	\vspace{-15pt}
	\caption{Comparison of attack success rates as we apply the diversity loss $L_{div}$ on the pixel/feature-level.
	All adversarial examples are crafted using IncRes-v2 as the surrogate model.  
	Best results are highlighted in \textbf{bold}.}
	\label{table:ap-tab2}
\end{table*}

\begin{table}[H]
	\begin{center}
        \resizebox{0.9\linewidth}{!}		
        {\begin{NiceTabular}{c|c|c|c}
        \hline
        Attack                      & Adv-Inc-v3                            & Ens-Adv-IncRes-v2                     & Ensemble  \\
		\hline
		\hline
		MI-FGSM                     & 28.3\%                                & 13.9\%                                & 15.4\%    \\
		DIM                         & 35.6\%                                & 22.0\%                                & 23.2\%    \\
        VMI-FGSM                    & 48.1\%                                & 38.1\%                                & 37.8\%    \\
        TAP                         & 30.8\%                                & 26.9\%                                & 28.0\%    \\
		FIA                         & 63.5\%                                & 34.4\%                                & 41.6\%    \\
		\cellcolor[gray]{0.85}Ours  & \cellcolor[gray]{0.85}\textbf{62.3\%} & \cellcolor[gray]{0.85}\textbf{38.7\%} & \cellcolor[gray]{0.85}\textbf{46.2\%} \\
        \hline              
        \end{NiceTabular}}
	\end{center}
	\vspace{-15pt}
	\caption{Comparison of attack success rates of different attacks on adversarial training models. 
	All adversarial examples are crafted using IncRes-v2 as the surrogate model. 
	Best results are highlighted in \textbf{bold}.}
	\label{table:ap-tab3}
\end{table}

\section{More Ablation Studies}
In this section, we provide additional ablation studies to verify the effectiveness of the individual components of our method.
For all experiments, we craft adversarial examples using IncRes-v2~\cite{inc-v4} as the surrogate model.

In Sec.~\ref{sec:methods}, we claim that our work differs from the work of Wu \etal~\cite{ata} on that we apply channel-wise normalization on the attention loss $L_{attn}$ (Eq.~\ref{eq:eq4}) to prevent perturbation on only the few feature channels with the highest magnitudes.
To verify this claim, we remove the normalization process from our framework and test its transferability, whose results are shown in Table~\ref{table:ap-tab1}.
Our method using channel-wise normalization shows higher attack success rates on all models than our method without channel-wise normalization, showing that it plays an important role perturbing the image attention in a diverse manner.

In DSGAN~\cite{dsgan}, the diversity loss is applied on a pixel- and feature-level.
We propose in Sec.~\ref{sec:methods} that our approach of enforcing diversity on the attention level to explicitly guide our attack generator to disrupt the \textit{salient} features in a diverse manner.
To verify this claim, we modify our diversity loss $L_{div}$ (Eq.~\ref{eq:eq5}) such that it maximizes the distance between the generated adversarial examples on the pixel- or the feature-level and report the attack success rates in Table~\ref{table:ap-tab2}.
Applying diversity on the feature-level leads to higher transferability than applying diversity on the pixel-level because it guides the generator to disrupt image features in a diverse manner.
However, na\"ively disrupting diverse features may not necessarily lead to diversity on the attention space, and our scheme of applying diversity on the attention space leads to higher transferability.

\section{Attack on Defense Models}
To compare the transferability of our method and the baseline methods on models with defense strategies, we test the attack success rates of each method on two adversarially trained models -- Adv-Inc-v3~\cite{at-scale} and Ens-IncRes-v2~\cite{ensemble}.
Table~\ref{table:ap-tab3} shows the attacks success rates of adversarial examples crafted using different attack methods on IncRes-v2~\cite{inc-v4} as the surrogate model.
Our method shows the highest ASR compared to all other methods, corroborating its high transferability even on models with defense techniques applied.